\documentclass[sigconf]{acmart}

\newcommand{\etal}{\textit{et al}.}
\usepackage{makecell}

\usepackage{multirow}
\usepackage{multicol}
\usepackage{arydshln}

\AtBeginDocument{%
	}

\copyrightyear{2023}
\acmYear{2023}
\setcopyright{rightsretained}
\acmConference[MM '23]{Proceedings of the 31st ACM International Conference on Multimedia}{October 29-November 3, 2023}{Ottawa, ON, Canada}
\acmBooktitle{Proceedings of the 31st ACM International Conference on Multimedia (MM '23), October 29-November 3, 2023, Ottawa, ON, Canada}
\acmDOI{10.1145/3581783.3612123}
\acmISBN{979-8-4007-0108-5/23/10}

\makeatletter
\gdef\@copyrightpermission{
	\begin{minipage}{0.3\columnwidth}
		\href{https://creativecommons.org/licenses/by-nc-sa/4.0/}{\includegraphics[width=0.90\textwidth]{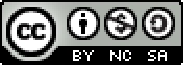}}
	\end{minipage}\hfill
	\begin{minipage}{0.7\columnwidth}
		\href{https://creativecommons.org/licenses/by-nc-sa/4.0/}{This work is licensed under a Creative Commons Attribution International 4.0 License.}
	\end{minipage}
	\vspace{5pt}
}
\makeatother

\begin{document}

%%
%% The "title" command has an optional parameter,
%% allowing the author to define a "short title" to be used in page headers.
%% \title[short title]{full title}
\title[MFR-Net: Multi-faceted Responsive Listening Head Generation via Denoising Diffusion Model]{MFR-Net: Multi-faceted Responsive Listening Head Generation via Denoising Diffusion Model}

%%
%% The "author" command and its associated commands are used to define
%% the authors and their affiliations.
%% Of note is the shared affiliation of the first two authors, and the
%% "authornote" and "authornotemark" commands
%% used to denote shared contribution to the research.
\author{Jin Liu}
\affiliation{
  \institution{Institute of Information Engineering, Chinese Academy of Sciences \& School of Cyber Security, University of Chinese Academy of Sciences}
  \streetaddress{}
  \city{Beijing}
  \country{China}
}
\email{liujin@iie.ac.cn}

\author{Xi Wang}
\authornote{Corresponding author.}
\affiliation{
	\institution{Institute of Information Engineering, Chinese Academy of Sciences}
	\streetaddress{}
	\city{Beijing}
	\country{China}
}
\email{wangxi1@iie.ac.cn}

\author{Xiaomeng Fu}
\affiliation{
	\institution{Institute of Information Engineering, Chinese Academy of Sciences \& School of Cyber Security, University of Chinese Academy of Sciences}
	\streetaddress{}
	\city{Beijing}
	\country{China}
}
\email{fuxiaomeng@iie.ac.cn}

\author{Yesheng Chai}
\affiliation{
	\institution{Institute of Information Engineering, Chinese Academy of Sciences \& School of Cyber Security, University of Chinese Academy of Sciences}
	\streetaddress{}
	\city{Beijing}
	\country{China}
}
\email{chaiyesheng@iie.ac.cn}

\author{Cai Yu}
\affiliation{
	\institution{Institute of Information Engineering, Chinese Academy of Sciences \& School of Cyber Security, University of Chinese Academy of Sciences}
	\streetaddress{}
	\city{Beijing}
	\country{China}
}
\email{caiyu@iie.ac.cn}

\author{Jiao Dai}
\affiliation{
	\institution{Institute of Information Engineering, Chinese Academy of Sciences}
	\streetaddress{}
	\city{Beijing}
	\country{China}
}
\email{daijiao@iie.ac.cn}

\author{Jizhong Han}
\affiliation{
	\institution{Institute of Information Engineering, Chinese Academy of Sciences}
	\streetaddress{}
	\city{Beijing}
	\country{China}
}
\email{hanjizhong@iie.ac.cn}

%%
%% By default, the full list of authors will be used in the page
%% headers. Often, this list is too long, and will overlap
%% other information printed in the page headers. This command allows
%% the author to define a more concise list
%% of authors' names for this purpose.
\renewcommand{\shortauthors}{Jin Liu et al.}

%%
%% The abstract is a short summary of the work to be presented in the
%% article.
\begin{abstract}
Face-to-face communication is a common scenario including roles of speakers and listeners. 
Most existing research methods focus on producing speaker videos, while the generation of listener heads remains largely overlooked.
Responsive listening head generation is an important task that aims to model face-to-face communication scenarios by generating a listener head video given a speaker video and a listener head image. 
An ideal generated responsive listening video should respond to the speaker with attitude or viewpoint expressing while maintaining diversity in interaction patterns and accuracy in listener identity information.  
To achieve this goal, we propose the \textbf{M}ulti-\textbf{F}aceted \textbf{R}esponsive Listening Head Generation Network (MFR-Net). Specifically, MFR-Net employs the probabilistic denoising diffusion model to predict diverse head pose and expression features. 
In order to perform multi-faceted response to the speaker video, while maintaining accurate listener identity preservation, we design the Feature Aggregation Module to boost listener identity features and fuse them with other speaker-related features. Finally, a renderer finetuned with identity consistency loss produces the final listening head videos. Our extensive experiments demonstrate that MFR-Net not only achieves multi-faceted responses in diversity and speaker identity information but also in attitude and viewpoint expression. 
  
\end{abstract}

%%
%% The code below is generated by the tool at http://dl.acm.org/ccs.cfm.
%% Please copy and paste the code instead of the example below.
%%



%%
%% Keywords. The author(s) should pick words that accurately describe
%% the work being presented. Separate the keywords with commas.

%%
%% This command processes the author and affiliation and title
%% information and builds the first part of the formatted document.
\maketitle

\begin{figure*}[t]
	\centering
	\includegraphics[width=0.9\textwidth]{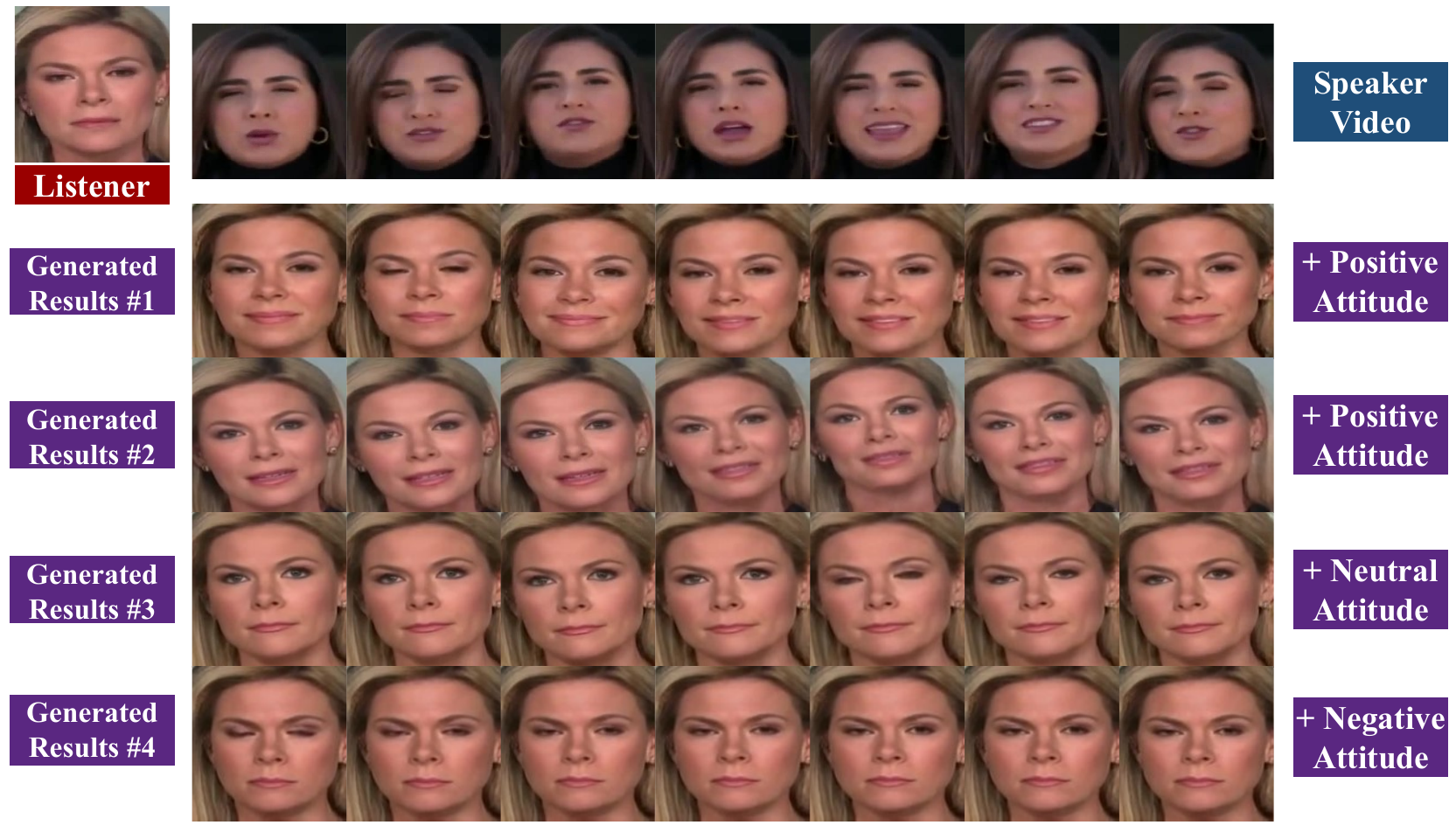} % Reduce the figure size so that it is slightly narrower than the column.
	\vspace{-3mm}
	\caption{Example results generated by MFR-Net. Given speaker video, audio, listener identity image and specific attitude label, our method generates natural multi-faceted responsive listening head videos. Results \#1 and \#2 display diverse results indicating positive attitude (smiling and nodding) while Results \#3 and \#4 show neutral (calm) and negative (serious) results. }
	\vspace{-2mm}
	\label{fig:teaser}
\end{figure*}

\section{Introduction} \label{sec:intro}

Face-to-face communication~\cite{drago2015effect,paul2016importance} is a common human activity, regularly occurring in various contexts such as between teachers and students, doctors and patients, and employers and employees. During such communication, participants take turns to act as speakers and listeners, engaging in an exchange of information. Typically, speakers articulate their thoughts verbally, while listeners convey their opinions non-verbally through head movements.

There exists extensive research works on speaker modeling, i.e. talking head generation~\cite{chen2020comprises, mirsky2021creation}, including improving the lip-syn quality~\cite{prajwal2020lip, park2022synctalkface, cheng2022videoretalking}, achieving free pose control~\cite{zhou2021pose, liang2022expressive, wang2022progressive, liu2023opt} or enhancing the naturalness by adding emotions~\cite{wang2020mead, sinha2022emotion, ji2021audio, ji2022eamm}. 
In face-to-face communication scenarios, the listener's role is also significant as that of the speaker. 
However, research on listener modeling remains less explored. 
Actually, the modeling of a listener is distinct from that of a speaker, as the former focuses on response to others while the latter primarily emphasizes lip synchronization.
The responsive listening head generation task aims to produce new listener head videos given the speaker talking head video and listener identity head image. Apart from modeling the daily scenario, the technique can also be used in digital avatar generation in Metaverse, customer representatives, robot communication, virtual audience modeling, wherever involves responsive listeners.

Listening is a conditioned action to reflect human behavior according to the principle proposed in social psychology and anthropology~\cite{barker1971listening}. The listener head videos should also contain the listener identity and opinion information at the same time. In light of the aforementioned background, there are multi-faceted requirements for the generation of responsive listener heads.
1) \textbf{Viewpoint Expression}: The generated responsive listener heads should convey certain viewpoints as a corresponding response to the speaker's head videos.Non-verbal behaviors such as nodding, smiling, frowning, head shaking, or neutral heads are generally used to convey those viewpoints.
2) \textbf{Speaker Interaction}: To achieve dynamic interaction between a speaker and listener in virtual communication, it is important that the pattern of listener motions exhibit a high correlation with the speaker's head video signal, as well as the speaker's attitude. The rhythm of the speaker's audio and the flow movement of their head can influence the action of the listener. Moreover, the attitude of the speaker, whether expressing agreement or disagreement, can prompt corresponding response actions in the listener, such as nodding, shaking, frowning, or other movements.
3) \textbf{Responsive Diversity}: For a given speaker video, listener identity, and attitude, there exist various natural response listener head videos. The expressive method and range of head movement may differ in every response scenario. In a virtual online meeting, it can be jarring if the avatars of each participant exhibit the same response head reactions, as this could detract from the sense of authenticity and spontaneity.
4) \textbf{Generation Naturalness}: The generated responsive listener heads should be of high image quality and free of any visual artifacts. Additionally, the identity information of the generated videos should match the given listener head image in order to ensure consistency and accuracy.

Recently, Zhou~\etal~\cite{zhou2022responsive} explore this task and collect the audio-visual ViCo dataset containing pairs of speaker and listener videos. To guide the generation of listening heads, they utilize an LSTM-based model to process the input signal and output listener head pose features. Later, {PCHG}~\cite{huang2022perceptual} post-process the generated frames with face segmentation model~\cite{Qin_2020_PR} to improve the stability of the background. However, the deterministic nature of their models fails to model the generation diversity, and the direct concatenation fusion between identity and speaker-related feature causes inaccurate identity preserving, leading to facial contour artifacts.

To tackle the aforementioned problems and meet the above multi-faceted requirements, we design the \textbf{M}ulti-\textbf{F}aceted \textbf{R}esponsive listening head generation network (MFR-Net).  Based on the denoising diffusion model~\cite{ho2020denoising}, MFR-Net is designed to predict the speaker's head pose and expression features. By leveraging the probabilistic nature of the denoising diffusion model, we manage to generate diverse listening head results.
Except for the generation diversity, the generated listening head videos should also interact with speakers with viewpoint or attitude expression and keep accurate listener identity information. To achieve the multi-faceted response, we propose the Feature Aggregation Module to embed the constraint conditions including speaker-related features, listener's attitude and listener identity information.
The proposed module is applied to each denoising and diffusion process to predict the noise term. In this way, multi-faceted constraints can be integrated into the diverse generated results.
Finally, the renderer trained with identity consistency loss is adopted to generate photo-realistic listener head images. 
As shown in Fig.~\ref{fig:teaser}, MFR-Net manages to achieve natural and diverse multi-faceted listening head generation with different listener attitudes and accurate listener identity preservation.

Our contributions are summarized as follows: 1) We propose MFR-Net, the first diffusion-based model for solving the task of generating responsive listener heads, and produces diverse and high-quality listener head videos. 2) The Feature Aggregation Module is designed to integrate speaker-related features, listener attitude and head image, leading to natural interaction with viewpoint expression and accurate listener identity information. 3) The state-of-the-art performance is achieved on the ViCo dataset in terms of visual quality, identity-preserving and generation diversity.

\section{Related work}

\subsection{Responsive Listening Head Generation}
Responsive Listening Head Generation aims to produce a head video of the listener, given a corresponding talking-head video of the speaker and face image of the listener. Zhou~\etal~\cite{zhou2022responsive} first propose this task and construct the ViCo dataset for evaluation. The proposed baseline utilizes LSTM to process the streaming input of visual and audio information of the speaker and produces facial 3DMM coefficients of the listener. %PIRenderer~\cite{ren2021pirenderer} is directly adopted as a generator to produce final head videos. 
Later, Huang~\etal~\cite{huang2022perceptual} utilizes pre-trained foreground-background segmentation model $\text{U}^{2}$Net~\cite{qin2020u2} to fuse and improve the background of generated results. However, the above method could merely generate solitary listening head videos given certain speaker talking videos, while MFR-Net manages to produce diverse listening head videos.

\begin{figure*}[t]
	\centering
	\includegraphics[width=0.95\textwidth]{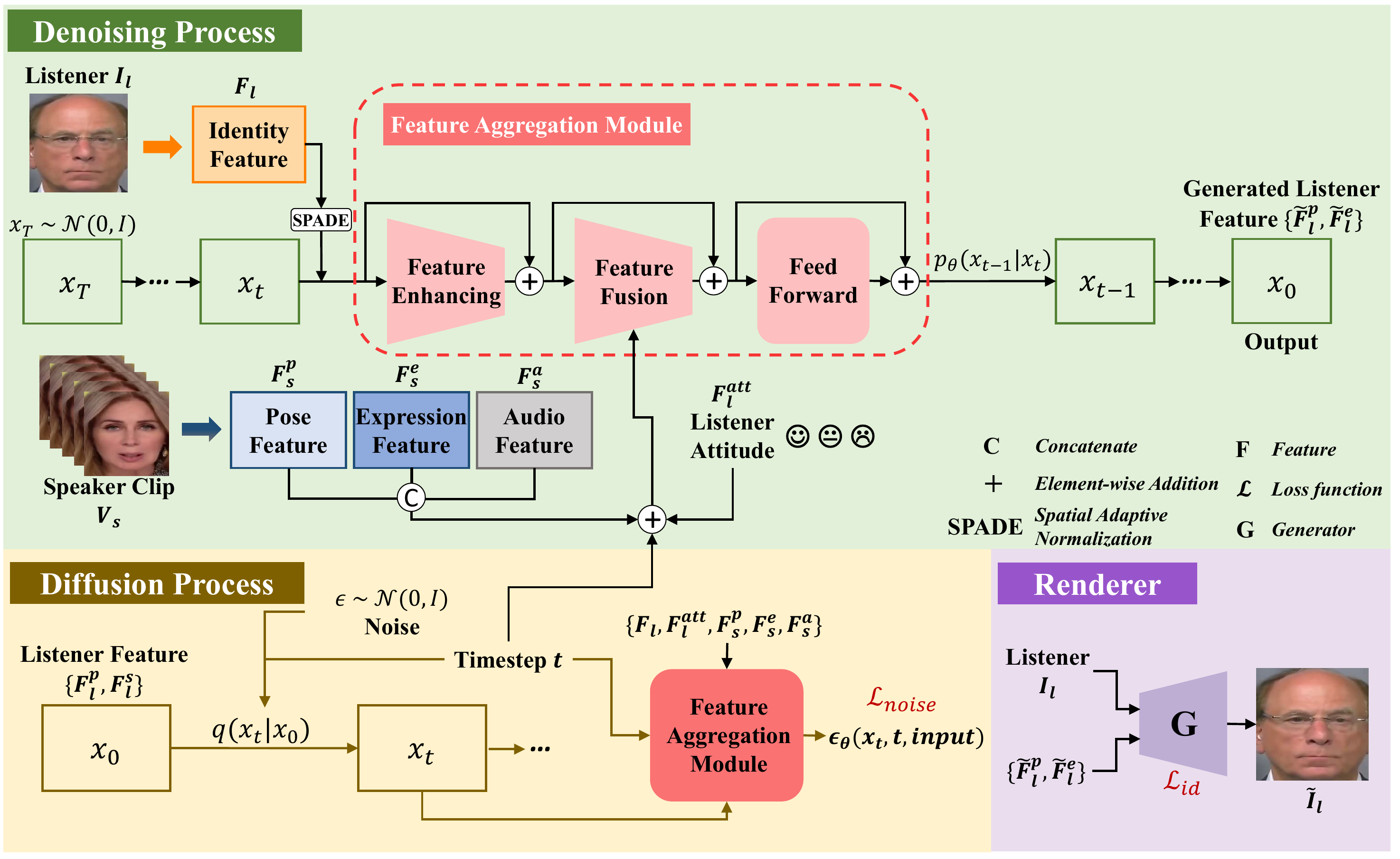} % Reduce the figure size so that it is slightly narrower than the column.
	\vspace{-3mm}
	\caption{\emph{Overview of MFR-Net.} MFR-Net adopts the denoising diffusion model to predict new listener features, which are fed into the generator along with listener image $I_{l}$ to produce responsive listening heads. The Feature Aggregation Module is designed to predict the noise used in the denoising diffusion model, given speaker clip $V_{s}$, listener image $I_{l}$, listener attitude $F_l^{att}$, intermediate feature $x_{t}$ and time step $t$.  The brown and black arrows indicate the training and inference process, respectively.}
	\vspace{-2mm}
	\label{fig:overview}
\end{figure*}

\subsection{Audio-Driven Head Synthesis}

Audio-driven head synthesis produces lip-sync talking head videos given source faces and driving speech signals. Some previous methods~\cite{fried2019text, zhang2021facial, lahiri2021lipsync3d, guo2021ad} generated talking head videos of specific identity used in the training process. Guo~\etal~\cite{guo2021ad} utilize NeRF-based~\cite{mildenhall2021nerf} network to model the head and torso separately and combine two generated parts. Zhang~\etal~\cite{zhang2021facial} perform 3D face reconstruction over each frame in the video and generate new expression coefficients to control mouth shape.  Though the aforementioned techniques preserve accurate source identity, they necessitate requiring high-quality videos of each subject for minutes to hours and only produce a limited range of identities, resulting in significant limitations on their applicability and generalization.

Therefore,  some recent methods~\cite{zhou2019talking,chen2019hierarchical,prajwal2020lip,chen2020talking, alghamdi2022talking} try to relieve the identity restriction and explore to generate any source subject. Prajwal~\etal~\cite{prajwal2020lip} adopt a pre-trained lip-sync discriminator to improve the lip-sync quality of generated talking head videos. Alghamdi~\etal~\cite{alghamdi2022talking} model each frame into the latent space of StyleGAN, map audio signals into displacements in the space and generate final talking images. Though the above methods have no identity mismatch problem since they merely edit the mouth areas, they generate unnatural talking heads given the still facial parts.

Later, to improve the diversity and naturalness of talking heads, current methods~\cite{zhou2021pose,zhang2021flow,liang2022expressive, wang2022one, zhang2022sadtalker,liu2023opt} explore to generate videos with head pose changes.  Some methods~\cite{zhou2021pose,liang2022expressive,liu2023opt} rely on auxiliary pose video to provide explicit guidance of pose sequences, while others~\cite{zhang2021flow,wang2022one,zhang2022sadtalker} explore to infer pose sequence from the audio signal. Among them, OPT~\cite{liu2023opt} try to solve the identity mismatch problem by disentangling identity and content feature from the audio signal to eliminate the effect of audio identity.  To improve the diversity of generated results, some works~\cite{zhang2022sadtalker, liu2023font} utilize the VAE structure to map the audio signal into diverse pose sequences. However, MFR-Net shows diversity not only in head poses, but also in facial expression parts as well.

\subsection{Diffusion Generative Models}
Denoising diffusion probabilistic model~\cite{ho2020denoising} is first proposed for unconditional image generation. Due to the stochastic property of the initialized noise in the reversion process, it manages to generate images of great diversity and soon becomes popular in a variety of creativity-oriented generation tasks. To name a few, GLIDE~\cite{nichol2021glide} introduces text conditions and proves the effectiveness of classifier-free guidance. DALLE-2~\cite{ramesh2022hierarchical} further generate semantics-consistent images conditioned on CLIP~\cite{radford2021learning} guidance. Later, Latent Diffusion Models~\cite{rombach2022high} perform the diffusion process in latent space rather than pixel space to improve the efficiency. The above works could generate diverse and natural images.  However, creative as they are, diffusion models adopted in such creativity-oriented tasks are not stable enough for responsible listening head generation, which aims to generate a natural video of a fixed listener while keeping identity information unchanged. Our proposed MFR-Net turns to generate facial coefficient features and impose explicit identity restrictions over the generation process to solve the above problems.

\section{Method}
Given speaker video clip $V_{s}$ containing visual and audio information, listener head image $I_{l}$ and attitude label, MFR-Net aims to produce multi-faceted responsive listening head videos. An overview of MFR-Net is shown in Fig.~\ref{fig:overview}, which contains four major components. The diffusion process and denoising process act as training and inference modules, respectively. The Feature Aggregation Module receives the inputs containing speaker video $V_s$, listener head image $I_l$, listener attitude and last intermediate state in the denoising or diffusion process, then adopts attention-based blocks to predict noise of current step $t$. The output of the denoising process is listener head pose and expression features, which will be fed into the generator to generate new listener head videos.

\subsection{Denoising Diffusion Model for Feature Generation}  \label{sec:ddpm}
Previous responsive listener head generation methods adopt deterministic LSTM-based modules to predict listener head features. They lack response diversity, which is the key factor in natural face-to-face interaction. To tackle the problem, we build our generation pipeline based on probabilistic denoising diffusion models.

The denoising diffusion model~\cite{ho2020denoising} is adopted to denoise a Gaussian noise step-by-step and finally generate the listener head pose and expression features. The form is as follows: $p_\theta\left(\mathrm{x}_0\right):=\int p_\theta\left(\mathrm{x}_{0: T}\right) d \mathrm{x}_{1: T},$
where $x_0$ is the real data of listener head pose and expression features and $x_1, ..., x_T$ are the latent data of the intermediate state. Specifically, the joint distribution  $p_\theta\left(\mathrm{x}_{0: T}\right)$ is called the denoising process, which is defined as a Markov chain with learned Gaussian transitions starting at $p\left(\mathbf{x}_T\right)=\mathcal{N}\left(\mathbf{x}_T ; \mathbf{0}, \mathbf{I}\right)$:

\begin{equation}
	\begin{aligned}
		& p_\theta\left(\mathbf{x}_{0: T}\right):=p\left(\mathbf{x}_T\right) \prod_{t=1}^T p_\theta\left(\mathbf{x}_{t-1} \mid \mathbf{x}_t\right) ,\\
		& p_\theta\left(\mathbf{x}_{t-1} \mid \mathbf{x}_t\right):=\mathcal{N}\left(\mathbf{x}_{t-1} ; \mu_\theta\left(\mathbf{x}_t, t\right), \Sigma_\theta\left(\mathbf{x}_t, t\right)\right).
	\end{aligned}
\label{equ:diff}
\end{equation}

Correspondingly, the diffusion process is the approximate posterior $q(x_{1:T}|x_0)$, which is also designed to fix a Markov chain that gradually adds Gaussian noise to the data according to a variance schedule  $\beta_1, ..., \beta_T$:

\begin{equation}
	\begin{aligned}
		& q\left(\mathbf{x}_{1: T} \mid \mathbf{x}_0\right):=\prod_{t=1}^T q\left(\mathbf{x}_t \mid \mathbf{x}_{t-1}\right), \\
		& q\left(\mathbf{x}_t \mid \mathbf{x}_{t-1}\right):=\mathcal{N}\left(\mathbf{x}_t ; \sqrt{1-\beta_t} \mathbf{x}_{t-1}, \beta_t \mathbf{I}\right).
	\end{aligned}
\end{equation}

As shown in the diffusion part in Fig.~\ref{fig:overview}, during training, noise $\epsilon$ and uniformly sampled time step $t$ are utilized to generate latent features. Given that the diffusion process admits sampling $x_t$ at an arbitrary time step $t$, instead of repeatedly adding noises to each intermediate state, the diffusion  process can be further denoted as $q\left(\mathrm{x}_t \mid \mathrm{x}_0\right)=\mathcal{N}\left(\mathrm{x}_t ; \sqrt{\bar{\alpha}_t} \mathrm{x}_0,\left(1-\bar{\alpha}_t\right) \mathbf{I}\right)$,
where $\alpha_{t}:= 1-\beta_{t}$ and $\bar{\alpha}_t:=\prod_{s=1}^t \alpha_s$.  To improve the performance, following iDDPM~\cite{nichol2021improved}, we choose to predict noise term $\epsilon$ instead of predicting latent feature. 
To integrate each input feature and achieve multi-faceted generation, we design the Feature Aggregation Module (Sec.~\ref{sec:agg}) instead of using traditional U-Net to predict the noise term. Hence, the loss term used to optimize the model parameters is as follows:
\begin{equation} \label{equ:loss}
	\mathcal{L}_{\text {noise }}=E_{t, x_0, \epsilon}\left[\left\|\epsilon-\epsilon_\theta\left(x_t, t, input\right)\right\|^2\right].
\end{equation}

Furthermore, as shown in Equation~\ref{equ:diff}, to generate new samples through the denoising process, $ \mu_\theta$ and $ \Sigma_\theta$ are demanded. Following Ho~\etal~\cite{ho2020denoising}, $ \Sigma_\theta$  is set to a constant number and $ \mu_\theta$ is:
\begin{equation}
	\mu_\theta\left(\mathbf{x}_t, t, input \right)=\frac{1}{\sqrt{\alpha_t}}\left(\mathbf{x}_t-\frac{1-\alpha_t}{\sqrt{1-\bar{\alpha}_t}} \epsilon_\theta\left(\mathbf{x}_t, t, input\right)\right).
\end{equation}

In this way, we manage to denoise the sampled noise step-by-step and finally generate new features, which are conditioned on the given input, as shown in the pink part in Fig.~\ref{fig:overview}.

\subsection{Feature Aggregation Module} \label{sec:agg}
In Sec.~\ref{sec:ddpm}, we show the denoising diffusion model as the listener feature generator, then we will illustrate the neural network for predicting the noise term $\epsilon_{\theta}(x_t, t, input)$. Specifically, the input includes identity features extracted from the face recognition model, pose and expression features reconstructed from 3DMM, audio features, listener attitudes and latent state features.
Unlike traditional denoising diffusion models~\cite{ho2020denoising, ramesh2022hierarchical, saharia2022photorealistic, rombach2022high} using the U-Net~\cite{ronneberger2015u} as basic module, to deal with input audio and pose features of variable length, we design our Feature Aggregation Module based on Transformer~\cite{vaswani2017attention} technique.

Furthermore, previous responsive listener head generation methods directly concatenate listener head features and other driving features, giving the same focus on features of different importance. In this way, responsive listener heads with inaccurate facial contour and the identity mismatch problem are generated. To alleviate the problem, we hope to enhance the speaker information into driving features by aggregating its most compatible identity features. 

Firstly, our objective is to enhance the input features and obtain a comprehensive understanding of the existing identity feature, which guides the generation of the noise term. Taking inspiration from the traditional face swapping work FaceShifter~\cite{li2019faceshifter}, we design our approach by incorporating the identity information into the latent feature map using the SPADE-like~\cite{park2019semantic} module. This integration ensures that the explicit identity information is embedded throughout the entire generation process, thereby effectively preserving the relevant information. Then we draw inspiration from the implementation of Transformer~\cite{vaswani2017attention} and utilize the multi-head self-attention module.
Specifically, details of this process can be formulated as follows:
\begin{equation} \label{equ:attention}
	\operatorname{Attention}(Q, K, V)=\operatorname{Softmax}\left(\frac{Q W_q\left(K W_k\right)^T}{\sqrt{C}}\right) V W_{v},
\end{equation}
where $W_q,W_k,W_v$ are projection parameters and $Q,K,V$ are the query, key, and value respectively.  During feature enhancing, we employ different projection weights on the combination of the identity feature and latent feature.

Secondly, to fusion the listener identity information into speaker driving features of another identity, we adopt the feature fusion module to build the correlation between input processed identity features and driving speaker features. The detailed formulation is the same as Equation~\ref{equ:attention} except that the key and value are from features from speaker clip $V_s$, listener attitude $F_l^{att}$ and time step $t$.

Finally, the feed-forward module is adopted to predict the noise term. Here we omit the residual connection and LayerNorm~\cite{ba2016layer} operation in each formula for simplification. The loss term in Equation~\ref{equ:loss} is utilized to train the module. Thanks to the Feature Aggregation Module, the listener identity information is enhanced and injected into each driving feature, thus helping to generate accurate listener features in the denoising process.

\subsection{Generator}
To improve the inter-frame coherence, the listener head pose features are obtained clip by clip with a fixed window and stride length through the above module. 
After obtaining the generated listener features, we adopt the state-of-the-art face reenactment model PIRenderer~\cite{ren2021pirenderer} to produce new listener heads. To further alleviate the identity mismatch problem, we re-train the model and add another identity restriction loss apart from the original perceptual loss, style loss and GAN loss.

\begin{equation}
	\mathcal{L}_{identity}=\left\| V (\tilde{I}_t)-V(I_t)\right\|_1, 
\end{equation}
where $V$ denotes the VGGFace~\cite{cao2018vggface2} model to extract identity features.  In this way, given generated listener pose and expression feature $\{\tilde{F}_{l}^{p}, \tilde{F}_{l}^{e}\}$ and listener head image $I_l$, new responsive listener head image $\tilde{I}_{l}$ is generated.

\begin{figure*}[t]
	\centering
	\includegraphics[width=0.9\textwidth]{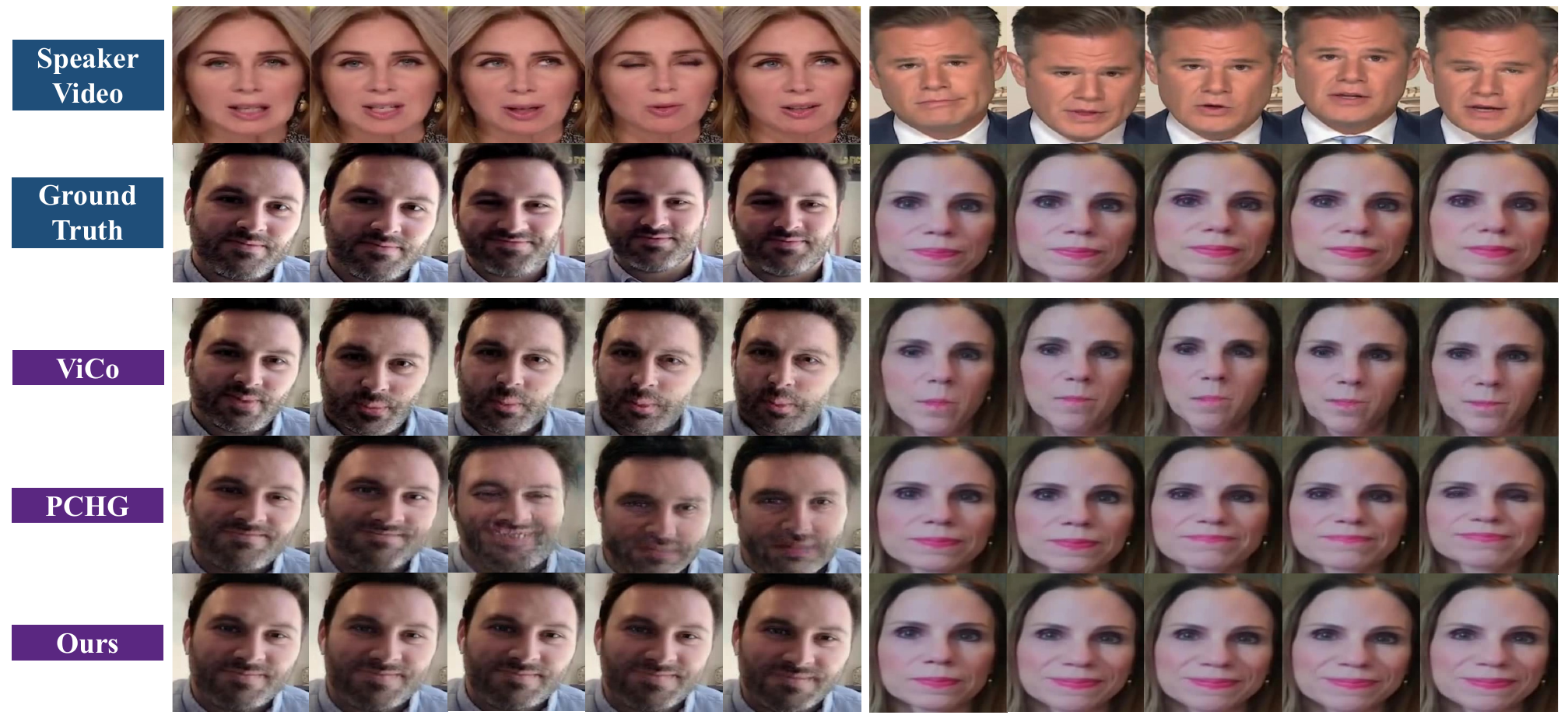} % Reduce the figure size so that it is slightly narrower than the column.
	\vspace{-3mm}
	\caption{Qualitative comparisons with other state-of-the-art methods conditioned by the same listener identity (the second row of each group) and the same attitude (left: positive listener in  $\mathcal{D}_{ood}$, right: positive listener in $\mathcal{D}_{test}$ ). }
	\vspace{-2mm}
	\label{fig:quality}
\end{figure*}

\begin{table*}[t]
	\centering
	\caption{The Feature Distance ($\times 100$) of different listening head generation methods. Each cell in the table represents the feature distance of { angle/expression/translation} coefficients respectively. Lower is better and the \textbf{bold} denotes the best result.}
	\vspace{-3mm}
	\label{tab:feature}
	\centering
	\begin{tabular}{ccccc:ccc:ccc:ccc}
		\toprule 
		\multirow{2}{*}{ Method } & \multirow{2}{*}{ Testset } & \multicolumn{3}{c:}{ Positive } & \multicolumn{3}{c:}{ Neutral } & \multicolumn{3}{c:}{ Negative} & \multicolumn{3}{c}{ Average } \\
		\cline { 3 - 14 } & & angle& exp& trans & angle& exp& trans & angle& exp& trans & angle& exp& trans  \\
		
		\midrule \multirow{2}{*}{ ViCO~\cite{zhou2022responsive}  } & $\mathcal{D}_{\text {test }}$ & 6.79 & 15.37 & 6.48 & 8.79 & 13.61 & 6.68 & 12.45 & 16.98&6.35&7.79&15.04&6.52 \\
		
		& $\mathcal{D}_{\text {ood }}$ & 9.72 & 24.89 & 9.51 & 6.33 & 23.51 & 8.95 & 8.54 & 18.99&5.81&8.23&22.83&8.32 \\
		
		\midrule \multirow{2}{*}{ PCHG~\cite{huang2022perceptual} } & $\mathcal{D}_{\text {test }}$ & 13.73 & 15.36 & 6.94 & 12.32 &15.35& 5.58 & 13.48 & 23.72&8.85&13.37&19.98&7.74 \\
		
		& $\mathcal{D}_{\text {ood }}$ & 17.93& 19.12 & 8.27 & 18.32 &17.66 & 9.69 & 17.47 & 20.86&17.95&18.00&18.82&8.76 \\
		
		\midrule \multirow{2}{*}{ Ours } & $\mathcal{D}_{\text {test }}$ & \textbf{5.36}& \textbf{13.73}&\textbf{5.94} &\textbf{ 5.35}& \textbf{12.32} & \textbf{4.58}& \textbf{11.78} & \textbf{13.46}&\textbf{5.48}&\textbf{6.82}&\textbf{13.37}&\textbf{6.02} \\
		
		& $\mathcal{D}_{\text {ood }}$ & \textbf{9.03}& \textbf{13.72} & \textbf{6.29} & \textbf{6.27}& \textbf{12.96 }&\textbf{4.77}  & \textbf{7.77} & \textbf{15.51}&\textbf{5.78}&\textbf{8.12}&\textbf{14.70}&\textbf{6.37} \\
		\bottomrule
	\end{tabular}
	\vspace{-2mm}
\end{table*}

\begin{table*}[t]
	\caption{{Quantitative comparisons on image-level metrics with state-of-the-art methods on ViCo dataset. The \textbf{bold} and \underline{underlined} notations represents the Top-2 results.   }}
	\vspace{-3mm}
	\centering
	\setlength\tabcolsep{6pt}
	
	\begin{tabular}{c c c c c c c}
		\toprule
		Method & SSIM $\uparrow$ & CPBD $\uparrow$ & PSNR$\uparrow$ & FID $\downarrow$ & CSIM  $\downarrow$ & Diversity  $\uparrow$   \\
		\midrule
		ViCO~\cite{zhou2022responsive}    &0.56   &  0.11&17.36    &27.74 & 0.23& 0.16\\
		PCHG~\cite{huang2022perceptual}    &{0.58}   &  {0.16}&{18.51}     &21.35 &0.25& 0.17 \\
		Ours    &   \textbf{0.59}& \textbf{0.18} & \underline{17.82}     & \textbf{20.08}& \textbf{0.18}& \textbf{0.28}\\
		\bottomrule
	\end{tabular}
	\vspace{-2mm}
	
	\label{tab:image}
\end{table*}

\section{Experiment}
\subsection{Experimental Settings}
\subsubsection{Dataset}
Our method is evaluated both quantitatively and qualitatively on the ViCo dataset~\cite{zhou2022responsive}, which is uniquely suited to our task. This dataset features 483 video clips capturing face-to-face interactions between two realistic subjects in a natural environment, with a total of over 0.1 million frames. Specifically, it includes the identities of 76 listeners and 67 speakers, and each response is manually annotated as positive, neutral, or negative attitude. As the only audio-visual dataset of its kind, ViCo provides an ideal benchmark for evaluating our approach.

\subsubsection{Comparison Methods}
Two responsive listener head generation methods are adopted as comparing methods:
\textbf{ViCo}~\cite{zhou2022responsive} adopts the LSTM-based Sequential Decoder to predict the pose and expression features of the listener subject and renders them into new responsive heads.
\textbf{PCHG}~\cite{huang2022perceptual} post-process the generated videos with a segmentation model to improve the stability of the background. 
Each method is trained on the training set $\mathcal{D}_{train}$ of ViCo, and further evaluated on the test set $\mathcal{D}_{test}$ and out-of-domain set $\mathcal{D}_{ood}$. Specifically, all identities in $\mathcal{D}_{test}$ have appeared in $\mathcal{D}_{train}$ while identities in $\mathcal{D}_{ood}$ have no overlap with ones in $\mathcal{D}_{train}$.

\subsubsection{Implementation Details}
The face video frames are cropped to $256\times256$ size at 30 FPS  and the audio signals are extracted into 45-dimensional acoustic features, including 14-dim mel-frequency cepstral coefficients (MFCC), 28-dim MFCC-Delta, energy, Zero Crossing Rate and loudness. The listener attitude information is denoted as the one-hot label. The window length of the speaker clip is set to be 40 frames with a sliding window length of 20 frames.
As for model details, we utilize multi-head attention with 8 heads and 4 layers. 
The identity feature is extracted from the face recognition model~\cite{cao2018vggface2}, while the pose and expression features are from the angles and translation coefficients of face 3DMM reconstruction operation~\cite{deng2019accurate} on the frames in each video. 
All experiments use an NVIDIA V100 GPU with 32 GB memory.

\subsection{Quantitative Evaluation}

\subsubsection{Evaluation Metrics}

In order to evaluate the precision of the generated speaker pose and expression features, we adopt the evaluation methodology utilized by ViCo~\cite{zhou2022responsive}, which involves measuring the $L_1$ distance between the generated features and their corresponding ground-truth features (FD). These features are extracted from 3D facial reconstruction data, where the angle and translation (trans) feature track changes in head pose, while the expression (exp) feature captures variations in facial movements.

To perform a comprehensive evaluation of the video-level performance, we adopt Fr'echet Inception Distance (FID)~\cite{heusel2017gans}, Structural Similarity (SSIM)~\cite{wang2004image}, Peak Signal-to-Noise Ratio (PSNR), and Cumulative Probability of Blur Detection (CPBD)~\cite{bohr2013no}. Additionally, to assess the quality of identity preservation, we utilize cosine similarity (CSIM) between identity features extracted from Vggface2~\cite{cao2018vggface2} on generated and ground truth images. Furthermore, in order to validate the diversity of the generated head motions, we compute the standard deviation of head motion feature embeddings extracted from the generated frames using Hopenet~\cite{ruiz2018fine}, which follows the methodology of SadTalker~\cite{zhang2022sadtalker}.

\subsubsection{Evaluation Results}
Table~\ref{tab:feature} presents the quantitative comparison results for both $\mathcal{D}_{test}$ and $\mathcal{D}_{ood}$ subsets of the ViCo dataset, with evaluations performed on head pose features including angle, expression, and translation.
It is worth noting that $\mathcal{D}_{test}$ shares listener identity overlap with the training set ($\mathcal{D}_{train}$), while $\mathcal{D}_{ood}$ does not. Results are presented for three different attitudes, as well as their average values.

MFR-Net achieves superior performance in head movements and facial expressions across all three attitudes.  On average evaluation using the $\mathcal{D}_{test}$ subset, angle, expression, and translation features show improvements of 0.97, 1.67, and 0.5 respectively. Notably, when generating unseen identities in $\mathcal{D}_{ood}$, the performance of other methods declines significantly compared to results obtained on  $\mathcal{D}_{test}$. In contrast, MFR-Net maintains relatively competitive performance, demonstrating its exceptional ability to generalize on unseen identities. These outstanding results are attributable to the proposed Feature Aggregation Module, which enhances and fuses identity information with speaker-related features.

Table~\ref{tab:image} presents image-level metrics for both $\mathcal{D}_{ood}$ and $\mathcal{D}_{test}$. The results demonstrate that MFR-Net outperforms other methods in terms of CSIM and Diversity scores, indicating superior preservation of identity information and generation of diverse images. Moreover, our method achieves competitive results with respect to image quality. Although PCHG achieves a slightly higher PSNR score due to its complicated post-processing on background pixels, MFR-Net generates images with accurate head pose features and natural-looking diverse interaction patterns.

\begin{table}[t]
	\caption{{User study results on ViCo dataset.   }}
	\vspace{-3mm}
	\centering
	\setlength\tabcolsep{4pt}
	\resizebox{\columnwidth}{!}{
		\begin{tabular}{c c c c c}
			\toprule
			Method & \makecell[c]{Overall  \\ Naturalness} & \makecell[c]{Motion \\ Diversity} &  \makecell[c]{Identity \\ Preserving} & \makecell[c]{Attitude \\ Matching} \\
			\midrule
			ViCO~\cite{zhou2022responsive}    &18.3\%   &  11.7\%&13.3\%    &29.7\% \\
			PCHG~\cite{huang2022perceptual}    & 36.6\%  &  25.7\%&25.0\%    &33.7\% \\
			Ours    & \textbf{49.0\% }  & \textbf{62.7\%} & \textbf{61.7\%}   &\textbf{36.7\%} \\
			\bottomrule
		\end{tabular}
	}
	\vspace{-3mm}
	\label{tab:user}
\end{table}

\subsection{Qualitative Evaluation}
In this section, we present the qualitative results from the generated responsive listening head frames using each method. The results are depicted in Fig.~\ref{fig:quality}. Our findings reveal that MFR-Net offers a reasonable response that may differ from the ground-truth, yet remains coherent. In comparison, other methods such as ViCo fail to maintain accurate facial contours and generate visible artifacts, while PCHG produces stiff response head videos that lack interactive features with the speaker video. Conversely, our approach ensures the preservation of accurate identity information with no visible artifacts. Additionally, the generated videos appear more visually plausible with natural and synchronized head motions and corresponding attitudes. For a detailed video-format comparison, please refer to the supplementary material.

\subsection{User Study}
We conduct user studies to evaluate the performance of all methods. We randomly choose 10 speaker videos and 10 listeners. For each cross-combination, three responsive videos of each attitude are generated. 
This process results in 300 generated videos for each method. To assess the quality, we asked 10 participants to watch the videos and choose the best method based on overall naturalness, motion diversity, identity preservation, and attitude matching quality. The results of the study are presented in Table~\ref{tab:user}. MFR-Net outperformed all other methods in all aspects, especially with regard to motion diversity and identity preservation. These findings indicate the superiority of our proposed Feature Aggregation Module and the effectiveness of our model.

\begin{figure}[t]
	\centering
	\includegraphics[width=\columnwidth]{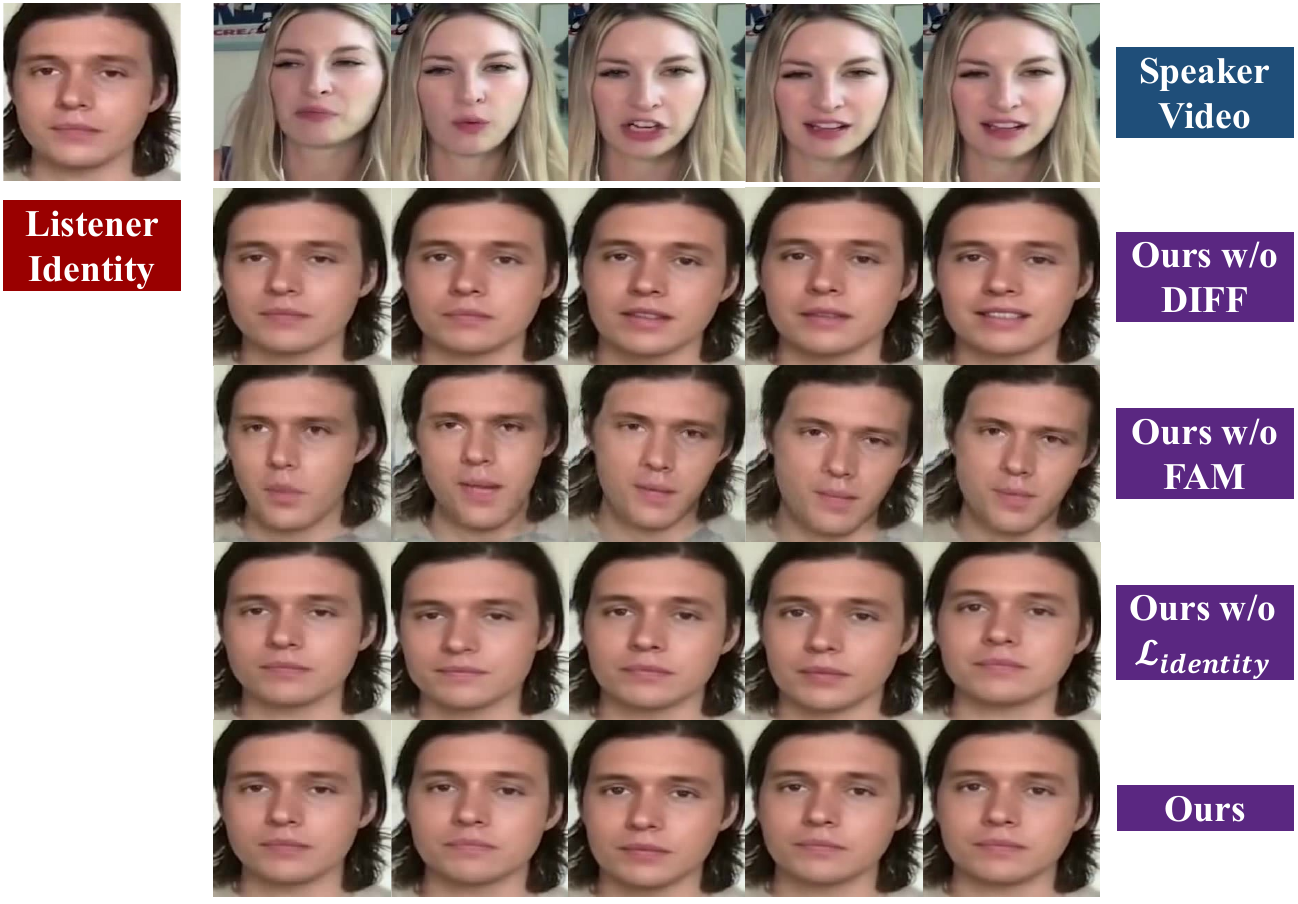} % Reduce the figure size so that it is slightly narrower than the column.
	\vspace{-7mm}
	\caption{ Qualitative results of ablation study on denoising diffusion model (DIFF), Feature Aggregation Module (FDM) and the identity consistency loss $\mathcal{L}_{identity}$.}
	\vspace{-2mm}
	\label{fig:ablation}
\end{figure}

\begin{table}[t]
	\caption{Ablation study for each proposed component in MFR-Net tested on both $\mathcal{D}_{ood}$ and $\mathcal{D}_{test}$ of ViCo dataset.}
	\vspace{-3mm}
	\label{tab:ablation}
	\centering
	\setlength\tabcolsep{3pt}
	\resizebox{\columnwidth}{!}{
		\begin{tabular}{c c c  c c c}
			\toprule
			\text { Method } & angle $\downarrow$  &  exp $\downarrow$ &  trans $\downarrow$ & CSIM $\downarrow$  & Diversity $\uparrow$ \\
			\midrule
			\text { w/o DIFF } & 7.48 & 14.89& 6.39& 0.21& 0.15\\
			\text { w/o FAM } & 7.52 & 18.27& 7.49& 0.27& 0.26\\
			\text { w/o $\mathcal{L}_{identity}$ } & - & - & - & 0.20& 0.28\\
			\text { Ours } & 7.47 & 14.04 & 6.20 & 0.18& 0.28\\
			\bottomrule
		\end{tabular}
	}
	\vspace{-3mm}
\end{table}

\subsection{Further Analysis}

\subsubsection{Ablation Study}
To assess the effectiveness of the designed components in MFR-Net,  we conduct the ablation study on the \emph{denoising diffusion model (DIFF)}, the \emph{Feature Aggregation Module (FAM)} and whether to utilize the \emph{identity loss $\mathcal{L}_{identity}$} in the renderer. 
Specifically, we compare our method without DIFF, which employs a simple LSTM-based model as the backbone for predicting noise terms, and our method without FAM, which concatenates all features directly and inputs them into the diffusion model. The results of both the qualitative and quantitative analyses are presented in Fig. \ref{fig:ablation} and Table \ref{tab:ablation}, respectively.
 
The experimental results demonstrate that incorporating the denoising diffusion model into MFR-Net leads to a significant improvement in generation diversity, owing to its probabilistic nature, while also ensuring the accurate synthesis of head features. Moreover, the Feature Aggregation Module and the identity consistency loss effectively preserve the precise listener identity information. The image-level results also attest to the effectiveness of our approach. Notably, when generating previously unseen identities in the out-of-distribution dataset $\mathcal{D}_{ood}$ , the FAM enhances the identity information and fuses it with other speaker-related features, resulting in natural-looking listening head videos. It is worth highlighting that this increased diversity in generated responses does not come at the cost of reduced accuracy in head feature synthesis.
 
\begin{figure}[t]
	\centering
	\includegraphics[width=\columnwidth]{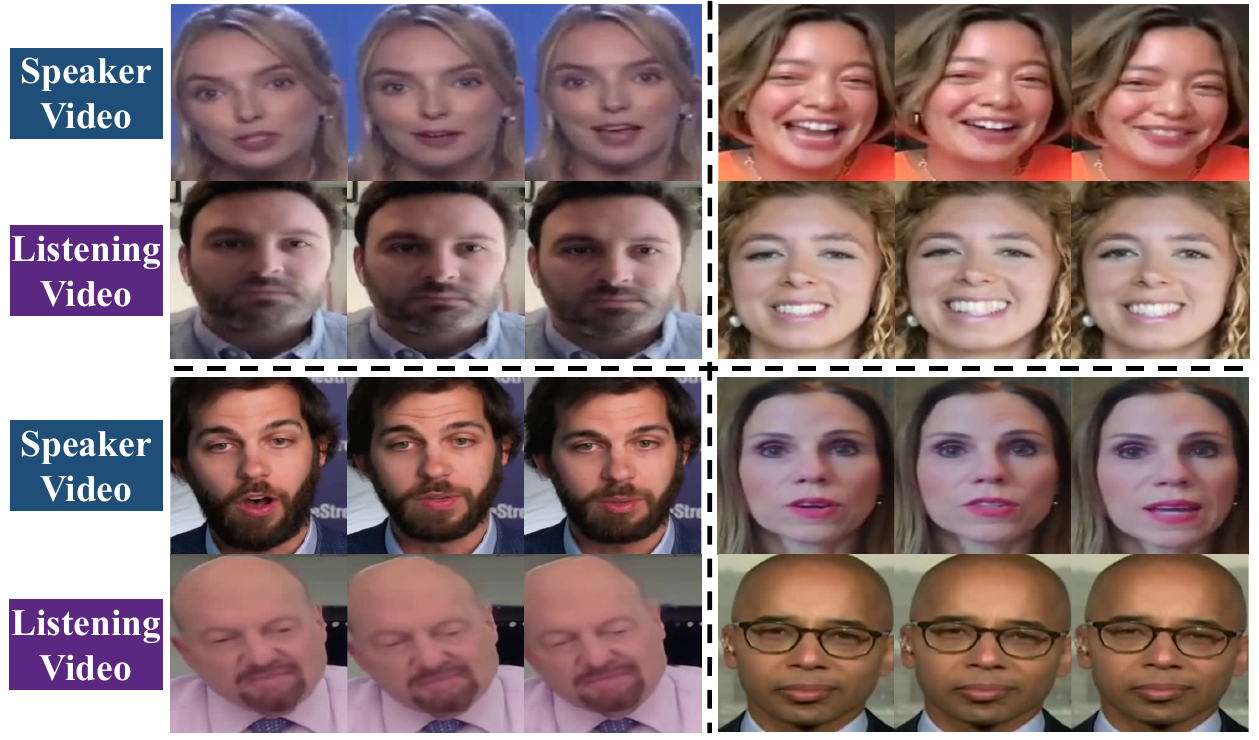} % Reduce the figure size so that it is slightly narrower than the column.
	\vspace{-6mm}
	\caption{ Visual interaction patterns in MFR-Net. The upper pairs show nodding and smiling for agreement, while the lower pairs display shaking for disagreement and a neutral attitude.}
	\vspace{-3mm}
	\label{fig:interaction}
\end{figure}

\begin{figure}[t]
	\centering
	\includegraphics[width=\columnwidth]{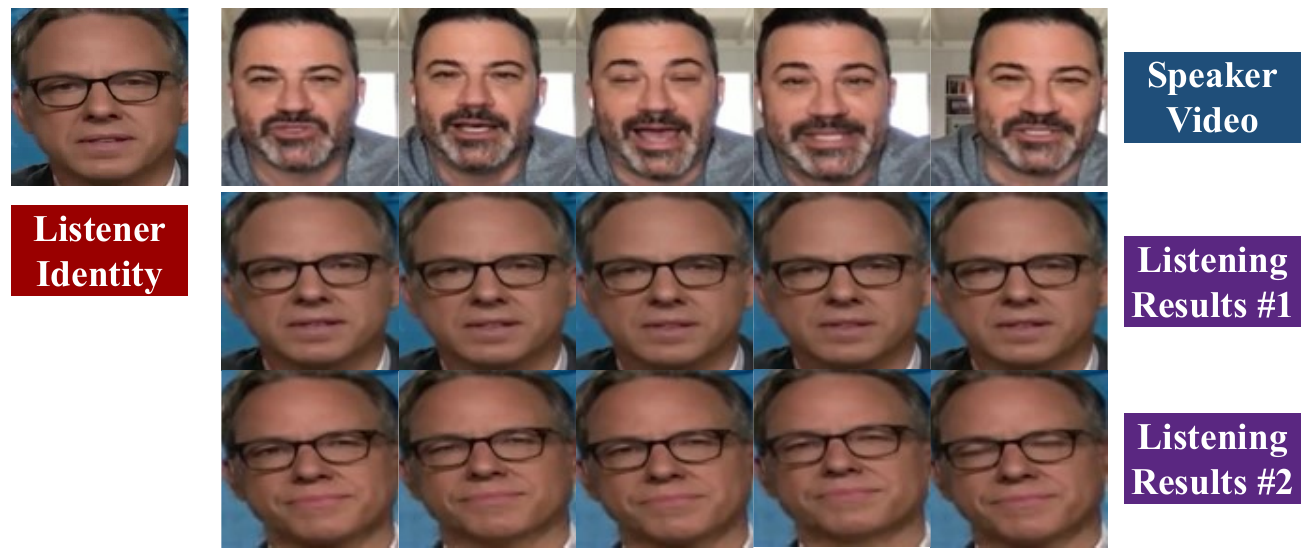} % Reduce the figure size so that it is slightly narrower than the column.
	\vspace{-6mm}
	\caption{ Visual results of diverse generation. To convey a negative attitude, frowning or serious emotion can be produced by MFR-Net.}
	\vspace{-3mm}
	\label{fig:diverse}
\end{figure}

\subsubsection{Interaction Patterns}
As highlighted in Section \ref{sec:intro}, generating responsive listening heads that can facilitate interaction is critical. To this end, we present several typical patterns generated by MFR-Net, as illustrated in Fig. \ref{fig:interaction}. In the upper pairs, we observe that when listeners intend to show agreement, they tend to nod their heads or smile. Similarly, the lower pairs demonstrate how MFR-Net can model disagreement by shaking heads or expressing neutrality through stationary head movements. These results indicate that MFR-Net is capable of generating diverse responses that can effectively facilitate interaction.

\subsubsection{Generation Diversity}
The denoising diffusion model within MFR-Net exhibits a probabilistic property that enables the generation of a diverse range of responsive listening heads, as illustrated in Fig.~\ref{fig:diverse}. Instead of producing entirely random output, MFR-Net generates varied results based on the listener identity image, speaker video, and attitude label. Depicted in Fig.~\ref{fig:diverse}, MFR-Net produces a frowning expression or a serious demeanor to convey negative attitudes. By leveraging this probabilistic mechanism, MFR-Net achieves greater flexibility and effectiveness in generating responses that accurately reflect the given conditions.

\begin{figure}[t]
	\centering
	\includegraphics[width=\columnwidth]{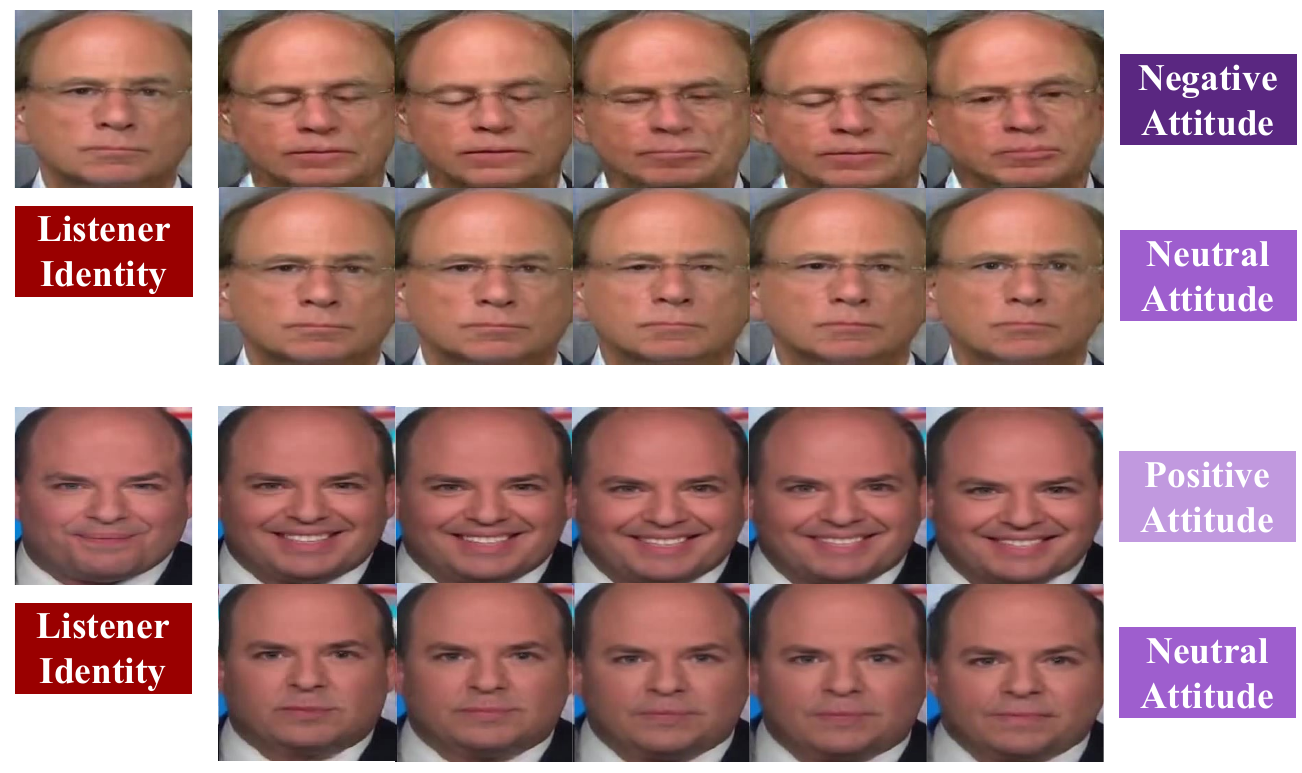} % Reduce the figure size so that it is slightly narrower than the column.
	\vspace{-6mm}
	\caption{ Visual results generated by MFR-Net conditioned by same speaker video and listener but different attitudes.}
	\vspace{-3mm}
	\label{fig:attitude}
\end{figure}

\subsubsection{Attitude Analysis}
To validate the response on attitude label, we show results generated by MFR-Net conditioned by the same speaker video and listener but different attitude labels, as shown in Fig.~\ref{fig:attitude}, where two distinct categories of output are displayed. Our analysis shows that the facial expressions and head motions generated by MFR-Net are highly expressive and distinguishable across various attitude labels, indicating the model's ability to produce multi-faceted responses. These findings contribute to a growing body of evidence supporting the efficacy of utilizing MFR-Net for generating natural responses with respect to human attitudes.

\subsection{Limitation}
While MFR-Net shows promising results in generating realistic and diverse listening head videos given a speaker video and listener head image, there are still some limitations that need to be addressed. One of the limitations is that 3DMMs do not model the variation of eyes and teeth, which may cause difficulties in synthesizing the finer details of teeth. Additionally, we only consider the speaker frames and audio signal without taking into account the semantic information contained in speech, such as speaker emotions and viewpoints. In future works, we plan to incorporate these factors to build a more realistic and highly interpretable system.

\section{Ethical Considerations}
MFR-Net is designed for modeling face-to-face communication scenarios and can be potentially utilized in world-positive use cases and applications, like digital avatar conversation and virtual online meetings. In case of misuse of the proposed method, we strongly support all relevant safeguarding measures against such malicious applications. 
We believe the proper usage of this technique will enhance the development of artificial intelligence research and relevant multimedia applications.

\section{Conclusion}
This paper proposes a novel method for generating responsive listening head videos. Our proposed MFR-Net utilizes the probabilistic denoising diffusion model to predict the listener head pose and expression features, thereby generating diverse and natural results. To achieve high-quality outputs with accurate response to speaker video while expressing certain attitude and preserving the listener identity, the Feature Aggregation Module is introduced, which enhances and fuses the multi-faceted responsive features. The proposed method is evaluated both quantitatively and qualitatively, and the experimental results demonstrate its superiority in generating precise and diverse listener head responses.

\section*{Acknowledgement}
This research is supported in part by the National Key Research and Development
Program of China (2020AAA0140000), and the National Natural
Science Foundation of China (No. 61702502).

%%
%% The next two lines define the bibliography style to be used, and
%% the bibliography file.
\bibliographystyle{ACM-Reference-Format}
\balance
\bibliography{sample-base}

%%
%% If your work has an appendix, this is the place to put it.
%\appendix
%
%\section{Research Methods}
%
%\subsection{Part One}

\end{document}